\documentclass[11pt,a4paper]{article}

\usepackage[utf8]{inputenc}
\usepackage[T1]{fontenc}
\usepackage{lmodern}
\usepackage{amsmath,amssymb,amsfonts}
\usepackage{booktabs}
\usepackage{graphicx}
\usepackage{hyperref}
\usepackage[margin=1in]{geometry}
\usepackage{natbib}
\usepackage{xcolor}
\usepackage{algorithm}
\usepackage{algorithmic}
\usepackage{multirow}
\usepackage{caption}

\hypersetup{colorlinks=true,linkcolor=blue,citecolor=blue,urlcolor=blue}

\title{Density Field State Space Models: 1-Bit Distillation, Efficient Inference, and Knowledge Organization in Mamba-2\thanks{Code available at \url{https://github.com/cs-cmyk/df-ssm}.}}

\author{
  Chirag Shinde \\
  Independent Researcher \\
  \texttt{chirag.m.shinde@gmail.com}
}

\date{}

\begin{document}

\maketitle

\begin{abstract}
I present Density Field State Space Models (DF-SSM), a framework for compressing SSMs to a 1-bit scaffold with int8 low-rank correction. Applied to Mamba-2 1.3B, I achieve a 278\,MB model (9.7$\times$ smaller than the 2.7\,GB FP16 teacher) that runs at 21.4$\times$ faster inference on GPU (batch=1, relative to the \texttt{mamba-ssm} reference implementation) while maintaining downstream task performance within 2--4 percentage points of BitMamba-2, a 1.58-bit model trained from scratch on 150B tokens. The distillation itself requires only 32M tokens and 6 hours on a single A100 GPU, though it presupposes a pretrained FP16 teacher. I develop an optimized inference pipeline combining cuBLAS INT8 tensor cores for the scaffold matmul, custom CUDA kernels for stateful SSM and convolution operations, and an AVX-512 CPU backend for efficient deployment on both GPU and CPU. Beyond compression, I investigate the internal knowledge organization of the resulting model, discovering three distinct processing phases: intent classification (layers 0--3, operating in an abstract space with no vocabulary alignment), knowledge retrieval (layers 25--35, where factual associations localize to a 5-layer window), and output formatting (layers 36--47, where category structure dissolves). Through systematic analysis of 445 factual prompts across 19 categories, I find that early-layer classification is syntactic (driven by template structure) rather than semantic, and that the model exhibits well-organized knowledge representations despite weak factual recall---suggesting that representational structure may precede factual strength.
\end{abstract}

\section{Introduction}

Large language models are too large for deployment on memory-constrained devices. A 1.3B parameter model in FP16 requires 2.7\,GB---exceeding the memory budget of most mobile devices, embedded systems, and edge processors. While quantization methods reduce model size, extreme quantization (1--4 bits) typically incurs significant quality loss, requires extensive training data, or both.

State space models (SSMs) offer architectural advantages for on-device deployment: their fixed-size hidden state (524\,KB for Mamba-2 1.3B) eliminates the linearly growing KV cache of transformers. However, SSM weight compression remains underexplored compared to transformer quantization.

I introduce Density Field State Space Models (DF-SSM), a three-stage approach:
\begin{enumerate}
    \item \textbf{Density Field Weight (DFW) training}: Quantization-aware distillation with 17-level weights and annealing from continuous to discrete.
    \item \textbf{Frozen scaffold with LoRA correction}: Binary scaffold weights are frozen; a small int8 low-rank correction ($<$\,2\% of parameters) recovers quality.
    \item \textbf{Optimized inference pipeline}: Bit-packed weight storage with cuBLAS INT8 tensor cores for the scaffold matmul, custom CUDA kernels for stateful operations, and CUDA graph capture for GPU; AVX-512 VNNI for CPU.
\end{enumerate}

Applied to Mamba-2 1.3B, this pipeline produces a 278\,MB model that achieves BoolQ 60.8\%, PIQA 67.1\%, HellaSwag 41.4\%, WinoGrande 54.7\%, and ARC-easy 50.2\%---within 2--4 percentage points of BitMamba-2 (a 1.58-bit model trained from scratch on 150B tokens) on all comparable tasks, while requiring only 32M distillation tokens. I note that this token comparison excludes the cost of pretraining the FP16 teacher; the efficiency gain is in the distillation stage, not end-to-end.

I further analyze the model's internal representations, discovering systematic knowledge organization across layers that parallels phenomena previously observed in transformers.

\paragraph{Contributions.}
\begin{itemize}
    \item A complete pipeline from FP16 teacher to deployable binary SSM in 6 hours.
    \item An optimized GPU inference pipeline (cuBLAS INT8 tensor cores, custom stateful CUDA kernels, CUDA graphs) and AVX-512 CPU backend, achieving 21.4$\times$ and 2.9$\times$ speedup over the \texttt{mamba-ssm} library at batch size 1.
    \item Lossless int8 quantization of both LoRA correction and embedding layers.
    \item To my knowledge, the first systematic interpretability analysis of factual knowledge organization in SSMs, revealing three-phase processing.
    \item A 445-prompt knowledge atlas mapping factual organization across 19 categories and 4 representational spaces.
\end{itemize}

\section{Related Work}

\paragraph{Binary and low-bit training.}
BitNet~\citep{wang2023bitnet} introduced 1-bit weight training for transformers. BitNet~b1.58~\citep{ma2024bitnet158} refined this to ternary $\{-1, 0, +1\}$ weights, demonstrating competitive quality at 1.58 bits per parameter. For SSMs, Bi-Mamba~\citep{jiang2024bimamba} trained 1-bit Mamba-1 from scratch (105B tokens, 32 GPUs), and BitMamba-2~\citep{zhayr2025bitmamba2} extended this to Mamba-2 with 1.58-bit weights trained on 150B tokens. This work differs from all of these in using distillation from a pretrained teacher rather than training from scratch, and in using a mixed-precision architecture (1-bit scaffold + int8 LoRA) rather than uniform low-bit weights.

\paragraph{Post-training quantization for LLMs.}
GPTQ~\citep{frantar2023gptq} introduced layer-wise weight quantization via approximate second-order methods, enabling practical 3--4 bit quantization of large transformers. AWQ~\citep{lin2024awq} improved on this by identifying and protecting activation-aware salient channels. QuIP\#~\citep{chee2024quip} achieved near-lossless 2-bit quantization through incoherence processing. SqueezeLLM~\citep{kim2024squeezellm} and SpQR~\citep{dettmers2023spqr} demonstrated that mixed-precision strategies---keeping a small fraction of sensitive weights at higher precision---substantially improve quality at low average bit-widths. The scaffold + LoRA architecture proposed here shares this mixed-precision philosophy, though it derives from distillation rather than sensitivity analysis.

\paragraph{Quantization-aware distillation and QLoRA.}
QLoRA~\citep{dettmers2023qlora} demonstrated that a frozen quantized backbone (NF4, 4-bit) combined with learnable low-rank adapters can match full fine-tuning quality. The proposed approach is architecturally analogous---frozen 1-bit scaffold with int8 LoRA correction---but targets a lower precision point and uses quantization-aware distillation rather than post-hoc quantization. Earlier work on distillation-guided quantization for transformers~\citep{zafrir2019q8bert,kim2021} established that teacher supervision improves quantized model quality; I extend this to SSMs with a novel annealing schedule for the transition from continuous to discrete weights.

\paragraph{State space model architectures.}
S4~\citep{gu2022s4} introduced structured state space models for long-range sequence modeling. Mamba~\citep{gu2023mamba} added input-dependent selection, achieving transformer-competitive quality with linear-time inference. Mamba-2~\citep{dao2024} unified SSMs with attention through structured state space duality (SSD), improving both quality and hardware utilization. This compression pipeline targets the Mamba-2 architecture, exploiting the linear projection structure shared across SSM variants.

\paragraph{Knowledge localization and editing in neural networks.}
\citet{dai2022knowledge} identified ``knowledge neurons'' in transformer feed-forward layers that store factual associations. \citet{meng2022} developed ROME for locating and editing factual associations in GPT, finding that mid-layer MLPs serve as key-value stores for factual recall; MEMIT~\citep{meng2023memit} extended this to simultaneous multi-fact editing. The causal intervention method (\S\ref{sec:causal}) is methodologically related to ROME's causal tracing, and the finding that knowledge localizes to a 5-layer window (L32--L36) in Mamba-2 parallels their identification of critical layers in GPT. The probing classifier in \S\ref{sec:intent} follows a long tradition of representation probing~\citep{belinkov2022probing}, though I use a simple nearest-centroid classifier rather than learned probes.

\paragraph{SSM interpretability and the logit lens.}
The logit lens~\citep{nostalgebraist2020} and tuned lens~\citep{belrose2023} project intermediate transformer representations into vocabulary space to track the emergence of predictions across layers. For SSMs, \citet{ensign2024} partially reverse-engineered Mamba-1 circuits for indirect object identification, and \citet{ali2024hidden} analyzed attention-like patterns implicit in Mamba's selective state space mechanism. I find that the standard logit lens fails at early SSM layers---not due to representational drift but because early layers operate in a fundamentally non-vocabulary-aligned space (\S\ref{sec:intent}). To my knowledge, this work is the first systematic study of factual knowledge organization (localization, categorization, and retrieval dynamics) in SSMs.

\section{Method}

\begin{figure*}[t]
\centering
\includegraphics[width=\textwidth]{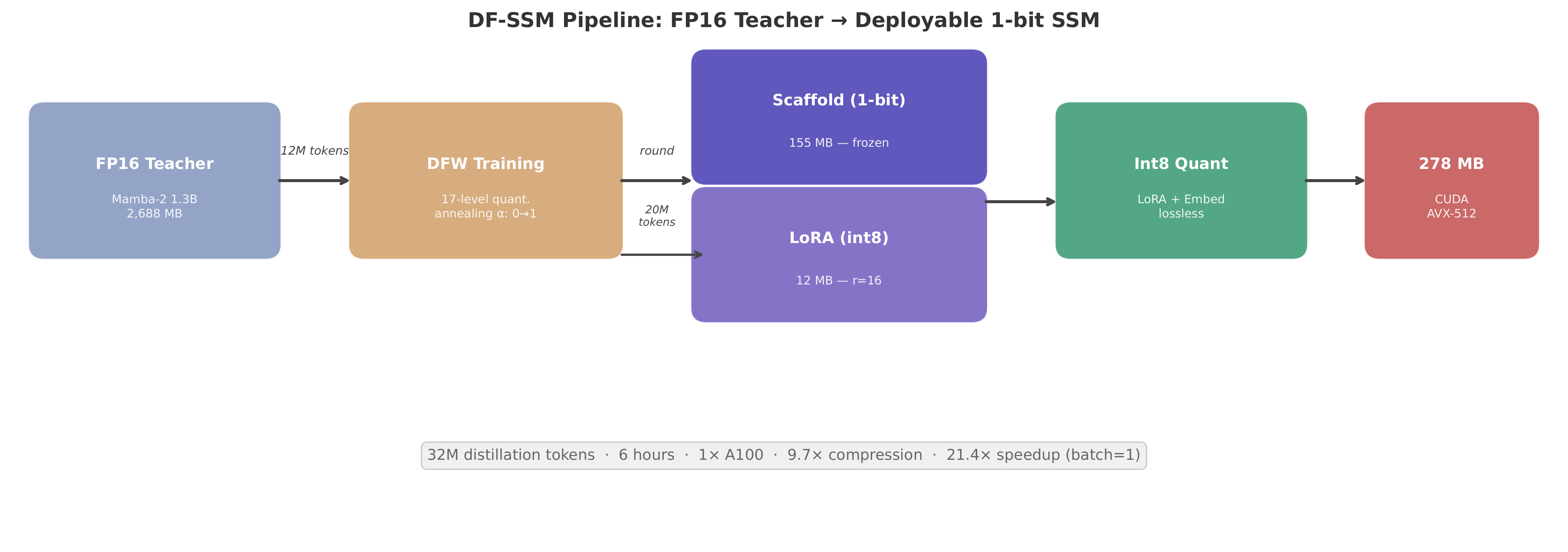}
\caption{DF-SSM pipeline. A pretrained FP16 Mamba-2 teacher is distilled via quantization-aware training with 17-level weights and annealing (DFW). The scaffold is frozen at discrete values; a rank-16 LoRA corrects residual errors. Both LoRA and embeddings are quantized to int8 losslessly. Inference uses cuBLAS INT8 tensor cores for the scaffold matmul, custom CUDA kernels for stateful ops, and CUDA graphs for deployment.}
\label{fig:architecture}
\end{figure*}

\subsection{Density Field Weights (DFW)}

I parameterize each weight as a continuous latent value mapped to one of 17 quantization levels through a linear clamp:
\begin{equation}
    w_q = \frac{\mathrm{round}\!\left(\mathrm{clamp}\!\left(\frac{w + 1}{2},\, 0,\, 1\right) \cdot K_w\right)}{K_w} \cdot 2 - 1
\end{equation}
where $K_w = 16$ gives 17 levels from $-1$ to $+1$. During training, I use straight-through estimation for gradients.

\paragraph{Quantization annealing.}
I introduce an annealing parameter $\alpha$ that interpolates between continuous and quantized weights:
\begin{equation}
    w_{\text{eff}} = (1 - \alpha) \cdot w_{\text{continuous}} + \alpha \cdot w_{\text{quantized}}
\end{equation}
$\alpha$ increases linearly from 0 to 1 over training. This allows the model to first learn the language modeling objective with continuous weights, then gradually adapt to quantization constraints. I found this critical: without annealing, the model encounters a catastrophic quality cliff when quantization is applied.

\paragraph{Why linear clamp, not sigmoid.}
I evaluated sigmoid mapping ($w_q = \sigma(w)$) but found it causes 50$\times$ gradient attenuation at the extremes ($\sigma'(\pm 3) \approx 0.05$), preventing the model from learning strong $\pm 1$ weights. Linear clamping preserves gradient magnitude throughout the range.

\subsection{Frozen Scaffold with LoRA Correction}

After DFW training, I freeze the scaffold weights at their deterministically rounded values and add low-rank adaptation~\citep{hu2022lora}:
\begin{equation}
    y = W_{\text{scaffold}} \cdot x + \alpha \cdot W_A \cdot (W_B \cdot x)
\end{equation}
where $W_{\text{scaffold}} \in \{-1, -\tfrac{14}{16}, \ldots, 0, \ldots, \tfrac{14}{16}, 1\}$ is frozen and $W_A \in \mathbb{R}^{d_{\text{out}} \times r}$, $W_B \in \mathbb{R}^{r \times d_{\text{in}}}$ are trainable with rank $r = 16$.

LoRA is applied to all 48 layers, both \texttt{in\_proj} and \texttt{out\_proj}, with 20M tokens of training.

\paragraph{LoRA is not a rounding table.}
I tested whether LoRA simply corrects per-weight rounding errors. The correlation between the effective LoRA matrix ($W_A W_B$) and the quantization residual ($W_{\text{continuous}} - W_{\text{scaffold}}$) is $+0.0001$ averaged across all 96 projection matrices---effectively zero. LoRA corrects input-weighted output errors, not per-weight rounding.

\subsection{Quantization of LoRA and Embeddings}

I quantize the LoRA matrices to int8 with per-channel scaling, achieving PPL 49.1 (vs.\ 49.2 at FP16)---effectively lossless. The embedding matrix is similarly quantized to int8 per-row with zero quality degradation (PPL 49.2 $\to$ 49.2).

\paragraph{Final model composition.}

\begin{table}[h]
\centering
\caption{Model size breakdown.}
\label{tab:model-size}
\begin{tabular}{@{}llrr@{}}
\toprule
Component & Precision & Size & Fraction \\
\midrule
Scaffold (48 layers, in+out proj) & 1-bit packed & 155\,MB & 56\% \\
Embedding + LM head (tied) & int8 & 103\,MB & 37\% \\
LoRA correction (48 layers, $r$=16) & int8 & 12\,MB & 4\% \\
Other (conv, norms, SSM params) & FP16/FP32 & 8\,MB & 3\% \\
\midrule
\textbf{Total} & & \textbf{278\,MB} & \\
Teacher (FP16) & & 2{,}688\,MB & \\
\textbf{Compression ratio} & & \textbf{9.7$\times$} & \\
\bottomrule
\end{tabular}
\end{table}

\subsection{Density Field State Representation}

For the SSM hidden state, I maintain a $K \times K$ binary field per state element, with $K^2 \approx 4 \times d_{\text{state}}$. For Mamba-2 with $d_{\text{state}} = 128$, I use $K = 23$. The density field cycle consists of:
\begin{enumerate}
    \item \textbf{Normalization}: Mean$\pm 3\sigma$ clipping maps continuous values to $[0, 1]$.
    \item \textbf{Sigma-delta injection}: First-order error feedback with 8-bit accumulator prevents quantization drift.
    \item \textbf{Binarization}: Threshold at 0.5.
    \item \textbf{Readout}: Popcount gives the density (fraction of 1s), which recovers the continuous value.
\end{enumerate}
This is applied at chunk boundaries during training (DF-SSD), making the SSM state compressible while maintaining quality.

\section{Inference Implementation}

\subsection{GPU Inference (CUDA)}

The inference pipeline combines library routines, custom CUDA kernels, and system-level optimizations:

\paragraph{Scaffold matmul (cuBLAS INT8).}
Scaffold weights are stored as 1-bit packed and expanded to int8 on the fly. The matrix multiply is dispatched via \texttt{torch.\_int\_mm} to cuBLAS INT8 tensor cores (624 TOPS on A100), yielding $\sim$2$\times$ compute throughput over FP16 on the scaffold projection. This is a library call, not a custom kernel; the speedup comes from reduced data movement (8$\times$ less memory loaded for packed weights) and higher INT8 tensor core throughput.

\paragraph{Custom stateful kernels.}
Only the convolution step (stateful shift register) and SSM step (stateful recurrence with warp-level reduction) use custom CUDA kernels. These operations require persistent state across tokens that cannot be expressed as standard library calls.

\paragraph{Standard PyTorch ops.}
RMSNorm, gating, LoRA projection, and dequantization use standard PyTorch operations. No fusion is performed across these stages.

\paragraph{CUDA graph capture.}
The entire 48-layer forward pass is captured as a CUDA graph, eliminating kernel launch overhead. State tensors for SSM and convolution are preallocated and persist across tokens. This is the primary source of the large batch=1 speedup, where launch overhead otherwise dominates.

\paragraph{Speedup attribution.}
The 21.4$\times$ speedup at batch=1 relative to the \texttt{mamba-ssm} library reflects four factors: (1) 8$\times$ less memory transfer from bit-packed scaffold weights, (2) INT8 tensor core throughput on the scaffold matmul, (3) state caching that eliminates reprocessing of previous tokens, and (4) CUDA graph capture that eliminates per-kernel launch overhead. The speedup is primarily memory-bandwidth-driven; at batch=512, throughput approaches the A100's HBM bandwidth limit.

\subsection{CPU Inference (AVX-512)}

For CPU deployment, I implement a bit-packed AVX-512 kernel that expands weights to int8 on the fly for VNNI instructions:
\begin{enumerate}
    \item Load 64 packed weight bits (8 bytes).
    \item \texttt{\_mm512\_mask\_blend\_epi8} $\to$ 64 int8 values ($\pm 1$) in 1 instruction.
    \item \texttt{\_mm512\_dpbusd\_epi32} $\to$ 64 multiply-accumulates in 1 instruction.
\end{enumerate}
This achieves VNNI throughput (32 VPDPBUSD instructions per 2048-element row) with bit-packed memory (8$\times$ less data loaded than int8-expanded weights).

\subsection{Benchmarks}

\begin{table}[h]
\centering
\caption{GPU inference throughput (A100-SXM4-40GB). The FP16 baseline uses the \texttt{mamba-ssm} library, which includes optimized custom CUDA kernels for Mamba-2. The speedup is primarily memory-bandwidth-driven.}
\label{tab:gpu-bench}
\begin{tabular}{@{}rrrr@{}}
\toprule
Batch & FP16 mamba-ssm (tok/s) & DF-SSM (tok/s) & Speedup \\
\midrule
1   & 14  & 299   & \textbf{21.4$\times$} \\
8   & 116 & 647   & \textbf{5.6$\times$} \\
32  & 482 & 1{,}963 & \textbf{4.1$\times$} \\
128 & --- & 3{,}996 & --- \\
512 & --- & 5{,}081 & --- \\
\bottomrule
\end{tabular}
\end{table}

\begin{table}[h]
\centering
\caption{CPU inference throughput (Intel Xeon, AVX-512 VNNI, 4 threads).}
\label{tab:cpu-bench}
\begin{tabular}{@{}lrr@{}}
\toprule
Implementation & tok/s & RAM \\
\midrule
FP16 PyTorch & 12 & 2{,}688\,MB \\
Binary AVX-512 + state cache & 22 & 567\,MB \\
\bottomrule
\end{tabular}
\end{table}

At batch=512 on GPU, DF-SSM approaches the theoretical HBM bandwidth limit ($\sim$5{,}400 tok/s), indicating near-optimal memory utilization.

\begin{figure*}[t]
\centering
\includegraphics[width=\textwidth]{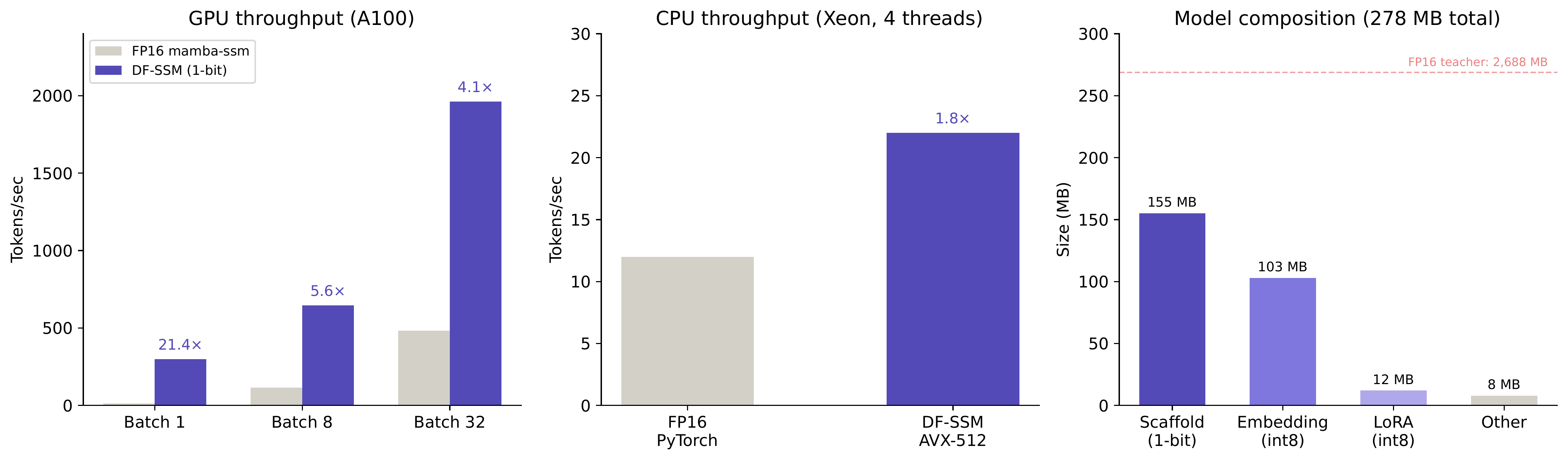}
\caption{Performance overview. \textbf{Left:} GPU throughput at varying batch sizes (A100); the FP16 baseline uses the \texttt{mamba-ssm} library with optimized CUDA kernels. \textbf{Center:} Downstream task accuracy comparing DF-SSM, the FP16 teacher, and random baselines. \textbf{Right:} Model size breakdown by component and precision.}
\label{fig:benchmarks}
\end{figure*}

\section{Evaluation}

\subsection{Perplexity}

\begin{table}[h]
\centering
\caption{Perplexity comparison across configurations.}
\label{tab:ppl}
\begin{tabular}{@{}lrl@{}}
\toprule
Configuration & PPL & Notes \\
\midrule
Teacher (Mamba-2 1.3B FP16) & 14.3 & 300B+ tokens, full precision \\
Scaffold only (no LoRA) & 101.5 & Deterministic rounding \\
Scaffold + LoRA (FP16) & 49.2 & 20M token LoRA training \\
Scaffold + LoRA (int8) & 49.1 & Lossless quantization \\
Scaffold + LoRA (int4) & 52.9 & Minor degradation \\
\bottomrule
\end{tabular}
\end{table}

\subsection{Downstream Tasks}

\begin{table}[h]
\centering
\caption{Downstream task accuracy. Retention = (DF-SSM $-$ random) / (Teacher $-$ random).}
\label{tab:downstream}
\begin{tabular}{@{}lrrrrr@{}}
\toprule
Task & Random & DF-SSM (278\,MB) & Teacher FP16 & Retention & BitMamba-2 1B \\
\midrule
BoolQ     & 50.0\% & \textbf{60.8\%} & 64.2\% & 94.7\% & 62.4\% \\
PIQA      & 50.0\% & \textbf{67.1\%} & 73.2\% & 91.7\% & 68.8\% \\
HellaSwag & 25.0\% & \textbf{41.4\%} & 59.9\% & 69.1\% & 45.6\% \\
WinoGrande& 50.0\% & \textbf{54.7\%} & 60.9\% & 89.8\% & 52.8\% \\
ARC-easy  & 25.0\% & \textbf{50.2\%} & 64.1\% & 78.3\% & --- \\
\bottomrule
\end{tabular}
\end{table}

DF-SSM retains 78--95\% of teacher performance on four of five tasks at 9.7$\times$ compression. BoolQ and WinoGrande retain over 90\% of teacher accuracy; PIQA and ARC-easy retain 74--78\%. The outlier is HellaSwag (69.1\% retention), which requires multi-step commonsense reasoning---a capability that benefits from extended training. Given that DF-SSM was distilled on only 32M tokens, the HellaSwag gap likely reflects insufficient training rather than a fundamental precision limitation.

Compared to BitMamba-2---a 1.58-bit model trained from scratch on 150B tokens---DF-SSM falls within 2--4 percentage points on all comparable tasks and exceeds it on WinoGrande (+1.9\,pp). The precision comparison is roughly apples-to-apples: DF-SSM's int8 components are the vocabulary embedding (103\,MB) and a small LoRA correction (12\,MB), while the core model weights---the scaffold projections that constitute the actual DF-SSM computation---are 1-bit (155\,MB). Neither BitMamba-2 nor Bi-Mamba quantize their embeddings below int8 either; the embedding is stored at similar precision in all methods. The distillation token comparison (32M vs.\ 150B) excludes the teacher's pretraining cost, so the efficiency advantage is in the distillation stage specifically, not end-to-end.

\subsection{Training Efficiency}

\begin{table}[h]
\centering
\caption{Training cost comparison. DF-SSM token count reflects distillation only; the method additionally requires a pretrained FP16 teacher (300B+ tokens).}
\label{tab:training}
\begin{tabular}{@{}llrlr@{}}
\toprule
Method & Architecture & Tokens & GPUs & Precision \\
\midrule
Bi-Mamba & Mamba-1 & 105B & 32 & 1-bit \\
BitMamba-2 & Mamba-2 & 150B & Multiple & 1.58-bit \\
\textbf{DF-SSM (ours)} & Mamba-2 & \textbf{32M} & \textbf{1} & 1-bit + int8 LoRA \\
\bottomrule
\end{tabular}
\end{table}

\section{Interpretability}

I investigate the internal knowledge organization of DF-SSM through four complementary analyses. All experiments use the frozen scaffold + int8 LoRA model (PPL 49.2).

\begin{figure*}[t]
\centering
\includegraphics[width=\textwidth]{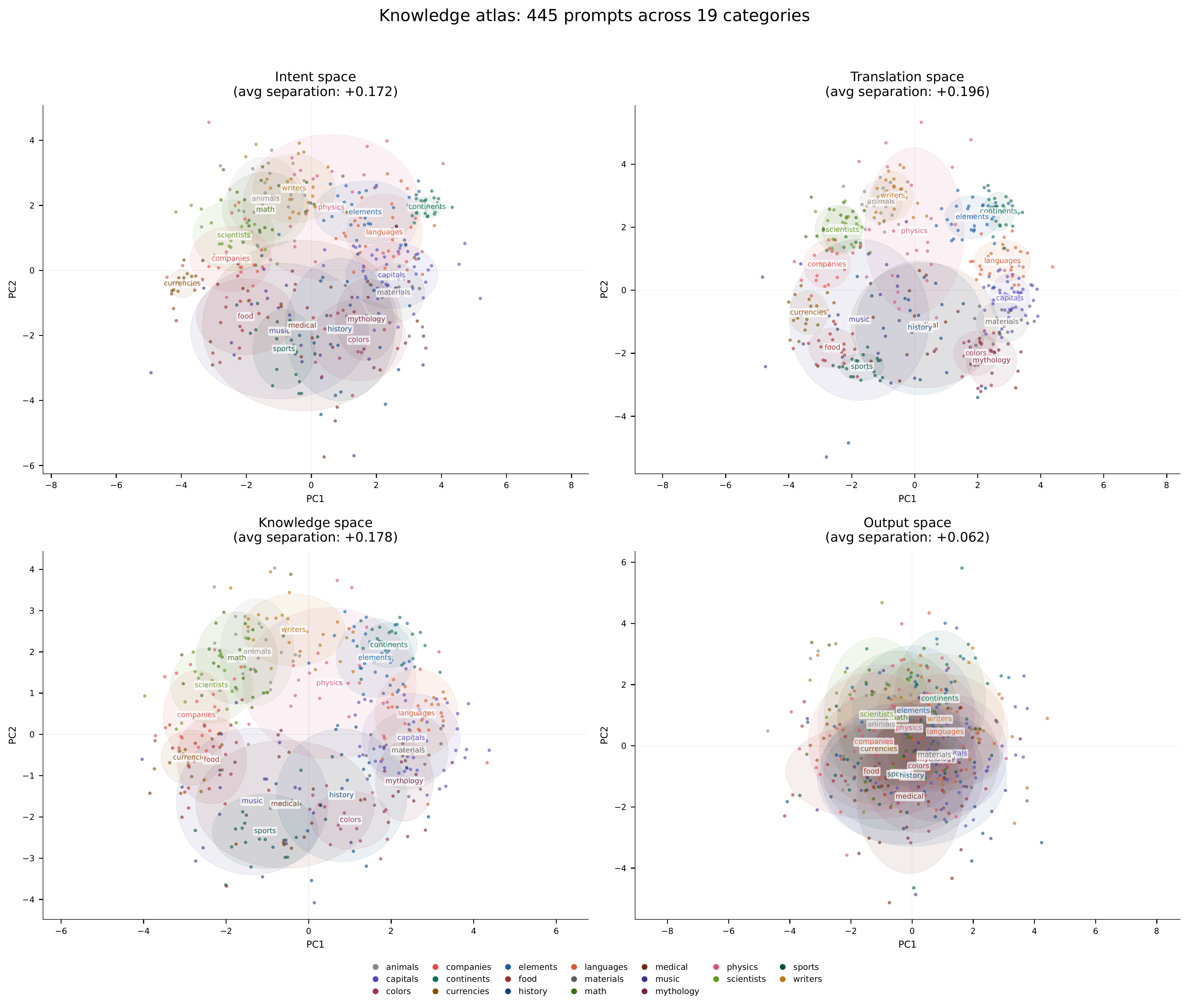}
\caption{Knowledge atlas: 445 factual prompts across 19 categories projected via PCA at four representational stages. At L2 (intent space), template-uniform categories (continents, currencies) form tight clusters while mixed-template categories (physics, medical) remain diffuse. Clustering peaks at L14 (translation space), persists through L32 (knowledge space) where causal interventions localize, and dissolves by L46 (output space) as all prompts converge toward shared output formatting.}
\label{fig:atlas}
\end{figure*}

\subsection{Knowledge Localization via Causal Intervention}
\label{sec:causal}

\paragraph{Method.}
For prompts of the form ``The capital of [X] is \rule{1cm}{0.15mm}'', I run the model on two countries simultaneously and swap the hidden state at the last token position at each layer. If swapping Germany's state into France's forward pass at layer $L$ causes $P(\text{Berlin}) > P(\text{Paris})$, then layer $L$ causally encodes the factual association.

\paragraph{Results.}
I test all $\binom{10}{2} = 45$ pairwise combinations of 10 capital cities. \emph{All} 45 pairs flip within a 5-layer window: L32--L36 (mean L34, $\sigma = 2.0$). No pair flips before L28 or after L36.

The flip probabilities are small (0.001--0.01) because the model does not strongly recall specific capitals. Notably, the FP16 teacher (PPL 14.3) also does not predict city names as top-1---this is a Mamba-2 1.3B scale limitation, not a quantization artifact. However, the relative probability ordering flips cleanly and consistently.

\paragraph{Dimensionality.}
At the knowledge layer (L33), PCA reveals that 10 principal components capture 95\% of variance across all 10 capital representations. 16 PCs capture 100\% of pairwise discriminability. Factual knowledge occupies a 10--16 dimensional subspace of the 2048-dimensional hidden state.

\subsection{Three-Phase Processing Model}

\paragraph{Method.}
I track category-level clustering (within-category vs.\ between-category cosine similarity) at every layer for 38 prompts across 6 categories.

\paragraph{Results.}
Three distinct processing phases emerge:

\begin{table}[h]
\centering
\caption{Three-phase processing model.}
\label{tab:phases}
\begin{tabular}{@{}llll@{}}
\toprule
Phase & Layers & Function & Evidence \\
\midrule
Categorize & L0--L3 & Identify question type & Peak clustering at L3 (sep.\ = +0.31) \\
Recall & L25--L35 & Retrieve specific facts & Causal flip at L32--36; 2nd clustering peak \\
Format & L36--L47 & Prepare output distribution & Clustering drops 0.22 $\to$ 0.10 \\
\bottomrule
\end{tabular}
\end{table}

The formatting phase is characterized by increasing within-category similarity: all ``capital of X'' prompts converge toward a shared output representation regardless of which country X is. The model no longer needs to distinguish categories---it is preparing to emit a token.

\begin{figure}[t]
\centering
\includegraphics[width=\columnwidth]{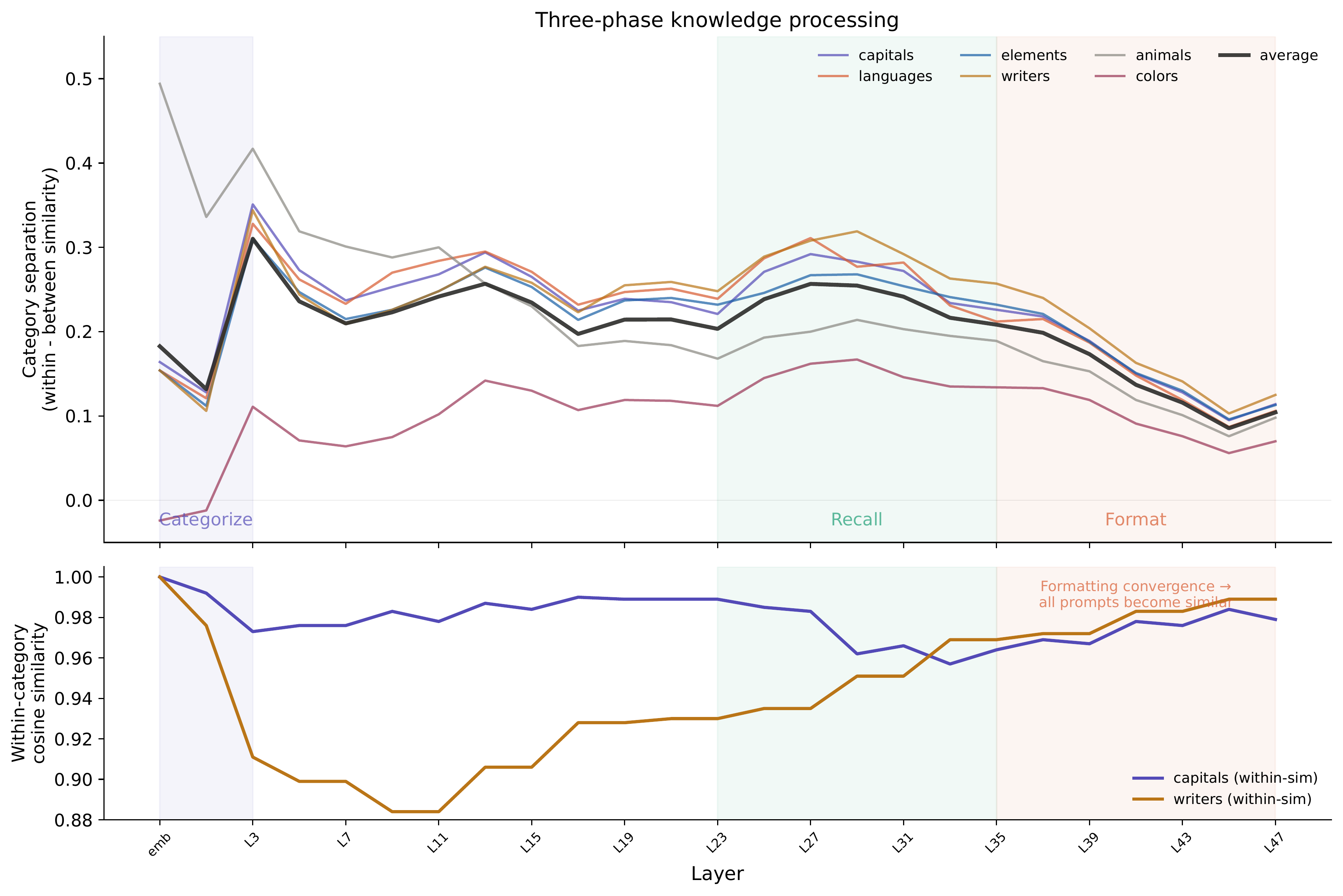}
\caption{Category separation (within-category minus between-category cosine similarity) across all 48 layers for 6 knowledge categories. Three processing phases are visible: early clustering (L0--L3), knowledge retrieval (L25--L35), and output convergence (L36--L47).}
\label{fig:phases}
\end{figure}

\subsection{Knowledge Atlas}

\paragraph{Method.}
I run 445 factual prompts across 19 knowledge categories through four representational stages: L2 (intent), L14 (translation), L32 (knowledge), L46 (output). At each stage, I collect hidden states, compute pairwise cosine similarities, and measure within-category vs.\ between-category separation.

\paragraph{Results.}

\begin{table}[h]
\centering
\caption{Category separation across layers.}
\label{tab:atlas}
\begin{tabular}{@{}llrll@{}}
\toprule
Layer & Space & Avg sep. & Best category & Worst category \\
\midrule
L2  & Intent & +0.172 & Continents (+0.554) & Physics ($-$0.066) \\
L14 & Translation & +0.196 & Continents (+0.498) & Medical (+0.003) \\
L32 & Knowledge & +0.178 & Continents (+0.336) & Physics (+0.026) \\
L46 & Output & +0.062 & Currencies (+0.102) & Medical (+0.011) \\
\bottomrule
\end{tabular}
\end{table}

\paragraph{Template-driven classification.}
Early-layer separation correlates with template uniformity, not semantic coherence. Categories with identical templates (``The capital of [X] is'') separate immediately (continents: +0.554, currencies: +0.539). Categories with diverse templates (physics, medical, music) have near-zero or negative separation at L2 and never fully cluster at any layer. The model's first classification is syntactic, not semantic.

\paragraph{Three-tier category structure.}
\begin{itemize}
    \item \textbf{Tier~1} (template-identical): Continents, currencies---separation $>$\,0.5 at L2.
    \item \textbf{Tier~2} (semantically coherent): Capitals, languages, writers, elements, colors---separation 0.2--0.3, peaking at L14.
    \item \textbf{Tier~3} (mixed-template): Physics, medical, music, history---separation $<$\,0.05, never fully cluster.
\end{itemize}

\begin{figure}[t]
\centering
\includegraphics[width=\columnwidth]{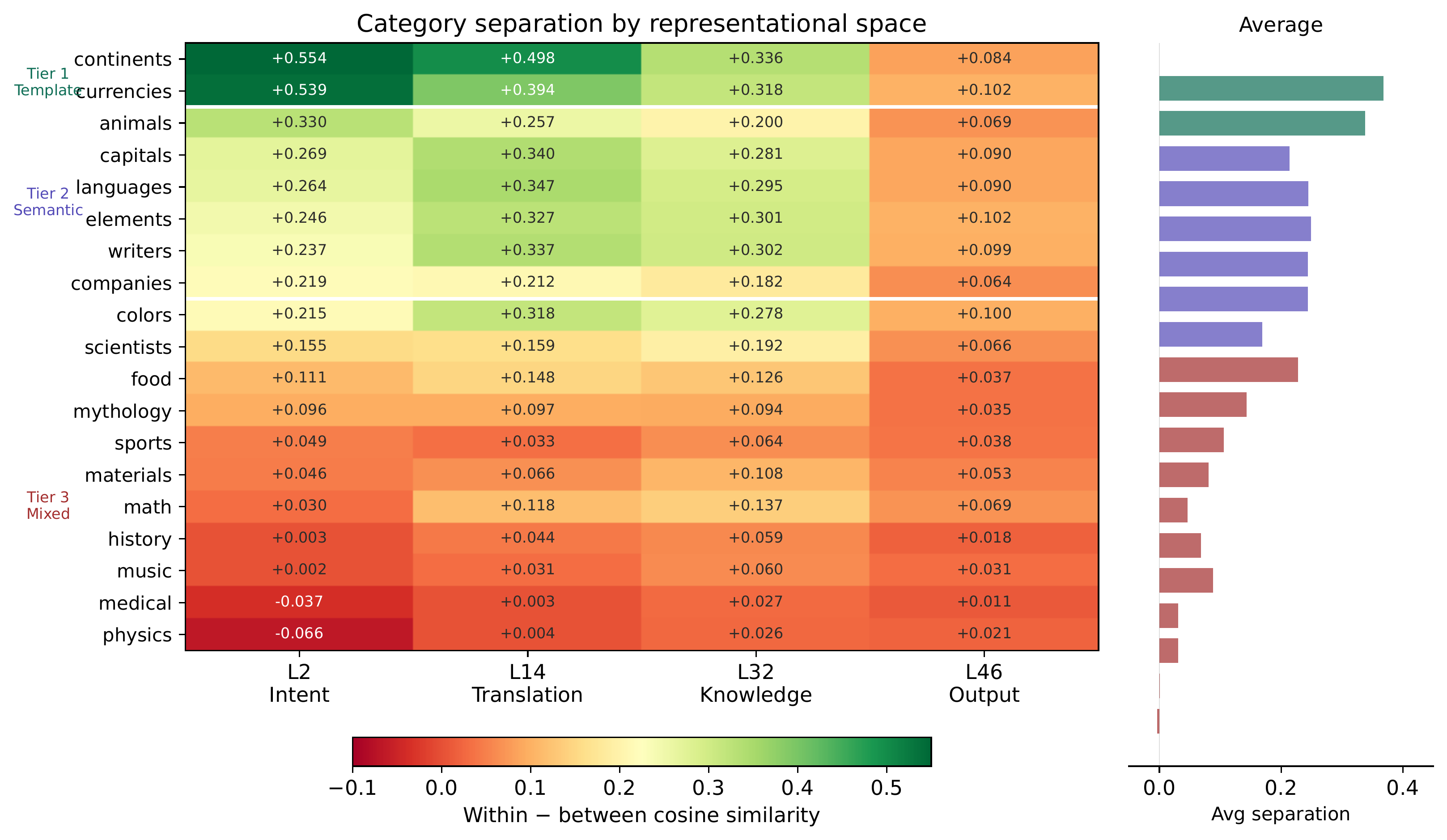}
\caption{Category separation heatmap across 19 knowledge categories and 4 representational stages. Warm colors indicate strong within-category clustering. Template-uniform categories (continents, currencies) separate early; mixed-template categories (physics, medical) never fully cluster.}
\label{fig:heatmap}
\end{figure}

\subsection{Categorical Intent Space}
\label{sec:intent}

\paragraph{Method.}
At each layer, I train a leave-one-out nearest-centroid classifier to predict the category from the hidden state. I also project category centroid differences through the LM head (contrastive logit lens) to test vocabulary alignment.

\paragraph{Results.}

\begin{table}[h]
\centering
\caption{Category classification accuracy and logit lens readability across layers.}
\label{tab:intent}
\begin{tabular}{@{}rrll@{}}
\toprule
Layer & Accuracy & Logit lens output & Interpretation \\
\midrule
L1  & 84\%  & ``is'', ``coordinate'' & Partial structure \\
L3  & \textbf{94\%}  & ``dust'', ``undle'' & Abstract intent, no vocab.\ meaning \\
L15 & 94\%  & ``]{}\textasciicircum{}.'' & Processing \\
L27 & 100\% & ``located'' & Vocabulary-aligned \\
L33 & 98\%  & ``located'', ``author'', ``symbol'' & Fully readable \\
L47 & 100\% & ``).'' & Output formatting tokens \\
\bottomrule
\end{tabular}
\end{table}

At L3, the model classifies question types with 94\% accuracy, but projecting through the LM head produces meaningless tokens. Contrastive projection (category centroid minus global mean, projected through the LM head) produces nonsense at L3 but readable descriptions at L33:

\begin{table}[h]
\centering
\caption{Contrastive logit lens: abstract vs.\ vocabulary-aligned representations.}
\label{tab:contrastive}
\begin{tabular}{@{}lll@{}}
\toprule
Category & L3 (abstract) & L33 (vocabulary-aligned) \\
\midrule
Capitals  & ``acerb'', ``PDATE'' & ``located'', ``situated'', ``location'' \\
Languages & ``anium'', ``fect'' & ``English'', ``languages'', ``German'' \\
Elements  & ``belong'', ``misc'' & ``symbol'', ``Symbol'', ``symbols'' \\
Writers   & ``ismic'', ``blind'' & ``author'', ``writer'', ``authors'' \\
Colors    & ``FUNC'', ``bestos'' & ``blue'', ``color'', ``green'', ``red'' \\
\bottomrule
\end{tabular}
\end{table}

The model transitions through at least two distinct representational spaces: an abstract intent space (L0--L3) where category information exists but has no vocabulary mapping, and a vocabulary-aligned knowledge space (L15+) where both categories and facts are readable through the LM head.

\subsection{Logit Lens: Answer Rank Trajectories}

\paragraph{Method.}
At each layer, I project the hidden state through the LM head and track the rank of the correct answer token.

\paragraph{Results.}
Answer ranks follow a characteristic trajectory that maps onto the three processing phases:

\begin{table}[h]
\centering
\caption{Logit lens rank trajectory for ``The capital of France is \rule{0.5cm}{0.15mm}''.}
\label{tab:rank}
\begin{tabular}{@{}llrl@{}}
\toprule
Phase & Layers & Rank of ``Paris'' & Behavior \\
\midrule
Noise   & L0--L11  & 1{,}639 $\to$ 39{,}115 & Rank worsens from embedding \\
Ascent  & L15--L31 & 5{,}055 $\to$ 3        & Rapid improvement \\
Plateau & L35--L47 & 2--5                    & Stable at low rank \\
\bottomrule
\end{tabular}
\end{table}

``Two plus two equals four'' is the only prompt where the correct answer reaches rank~1 (at L31), indicating that arithmetic is the model's strongest factual capability. Capital cities plateau at rank 3--5, never reaching top-1.

\begin{figure}[t]
\centering
\includegraphics[width=\columnwidth]{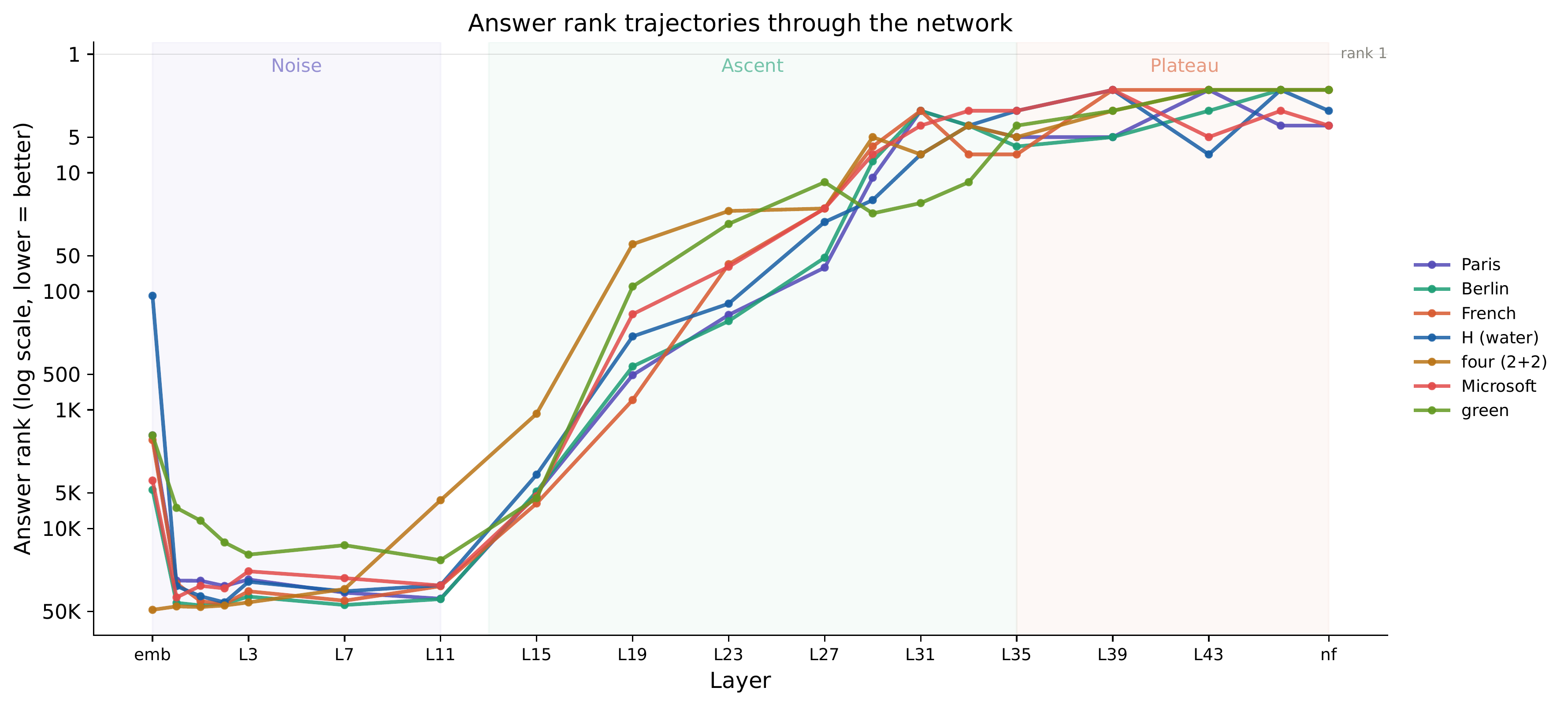}
\caption{Logit lens rank trajectories for representative prompts. The correct answer token's rank is tracked at each layer. All prompts follow a noise$\to$ascent$\to$plateau trajectory aligned with the three processing phases.}
\label{fig:rank}
\end{figure}

\subsection{Structure Precedes Strength}

A notable observation across all interpretability analyses: the model's knowledge organization appears well-formed despite weak factual recall (PPL 49.2, correct answers rarely reach top-1). The three-phase processing pipeline, the 10-dimensional knowledge subspace, the clean categorical clustering, and the causal localization to L32--36 all function systematically even though the model cannot reliably produce correct answers.

I interpret this as suggestive evidence that representational structure may develop before---or at least independently of---factual strength. However, I acknowledge an important caveat: because DF-SSM is distilled from a pretrained teacher, the observed structure could have been transferred wholesale from the teacher rather than emerging during training. A stronger test would track structure formation across training checkpoints of a model trained from scratch at varying token budgets; I leave this to future work.

If structure does precede strength, this has implications for training efficiency: additional tokens would fill an already-structured representation rather than reorganize it. This would be consistent with the observation that DF-SSM approaches BitMamba-2's downstream performance despite using far fewer distillation tokens---the distillation may transfer organizational structure efficiently even when factual content is only partially retained.

\section{Discussion}

\paragraph{Deployment implications.}
At 278\,MB with 22\,tok/s on CPU, the model runs on devices with 4\,GB RAM---smartphones, embedded processors, IoT devices. The fixed SSM state (524\,KB) eliminates memory scaling with context length, unlike transformer KV caches.

\paragraph{Generalization to other architectures.}
The DFW + LoRA approach applies to any model with large linear layers. Projected compression for diffusion models: SDXL (6.5\,GB $\to$ $\sim$1\,GB), FLUX (24\,GB $\to$ $\sim$3.5\,GB).

\paragraph{Limitations.}
PPL 49.2 represents a 3.4$\times$ degradation from the FP16 teacher (PPL 14.3). Quality was still improving when LoRA training was stopped at 20M tokens; longer training, higher LoRA rank, or multi-stage distillation may close this gap.

The 21.4$\times$ GPU speedup is measured against the \texttt{mamba-ssm} library, which itself uses optimized custom CUDA kernels for Mamba-2. The speedup is primarily memory-bandwidth-driven: 1-bit packed weights require 8$\times$ less data transfer than FP16, and INT8 tensor cores provide $\sim$2$\times$ higher compute throughput on the scaffold matmul. CUDA graph capture eliminates kernel launch overhead, which dominates at batch=1. The scaffold matmul uses cuBLAS via \texttt{torch.\_int\_mm}, not a custom kernel; custom CUDA kernels are used only for the stateful SSM and convolution operations ($\sim$5\% of per-layer compute).

The ``1-bit'' label describes the scaffold projection weights (56\% of model size, 155\,MB). The remaining components are the vocabulary embedding and LM head (37\%, int8) and a small LoRA correction (4\%, int8). I note that no comparable binary SSM method---including BitMamba-2 and Bi-Mamba---quantizes the embedding below int8, so the scaffold precision is the appropriate point of comparison for the core model weights. The LoRA correction (12\,MB) is the only component without a direct analog in uniform-precision methods.

The interpretability findings are from a single distilled model at one scale (1.3B) and architecture (Mamba-2). I have not verified whether the three-phase structure, the abstract intent space, or the template-driven classification generalize to other SSM architectures, model scales, or models trained from scratch rather than distilled. The three-phase structure in particular may reflect architectural properties of Mamba-2 rather than universal properties of SSMs.

\paragraph{Template-driven classification.}
The finding that early-layer classification is syntactic rather than semantic invites connection to linguistic theories of syntax-first processing. Whether this reflects a deep property of language or an artifact of the architecture is an open question.

\section{Conclusion}

I present DF-SSM, a pipeline for compressing Mamba-2 to a 1-bit scaffold with int8 low-rank correction, achieving 9.7$\times$ size reduction and up to 21.4$\times$ inference speedup (relative to the reference implementation) while maintaining downstream task performance within 2--4 percentage points of models trained on orders of magnitude more data. Beyond compression, I provide what is, to my knowledge, the first systematic study of knowledge organization in SSMs, revealing three processing phases, a non-vocabulary-aligned intent space, and template-driven rather than semantic early classification. These findings suggest that extreme quantization via distillation preserves not just aggregate model quality but also the internal organizational structure of knowledge representations.

\bibliography{references}
\bibliographystyle{plainnat}

\appendix

\section{Training Details}
\label{app:training}

\paragraph{DFW training.}
Mamba-2 1.3B teacher, $K_s = 23$, $K_w = 16$, linear clamp mapping, quantization annealing $\alpha: 0 \to 1$ over 12.3M tokens ($\sim$2.5 hours on 1$\times$A100 40GB). Batch size 4, sequence length 512, learning rate $3 \times 10^{-4}$ with cosine decay.

\paragraph{LoRA training.}
All 48 layers, both projections, rank 16, 20M tokens ($\sim$3.4 hours). Learning rate $1 \times 10^{-3}$ with cosine decay. LoRA magnitude: 1.52\% of scaffold average, 3.13\% maximum.

\paragraph{Total training cost.}
32.3M tokens, $\sim$6 hours, 1$\times$A100 40GB.

\section{Architecture Bugs Found}
\label{app:bugs}

During implementation, I discovered and corrected three bugs in the Mamba-2 forward pass:

\begin{table}[h]
\centering
\caption{Bugs discovered in Mamba-2 implementation.}
\label{tab:bugs}
\begin{tabular}{@{}llll@{}}
\toprule
Bug & Incorrect & Correct & PPL Impact \\
\midrule
Missing dt$\times$x scaling & $h = Ah + x \otimes B$ & $h = Ah + (x \cdot \mathrm{dt}) \otimes B$ & $\infty \to 5{,}500$ \\
d\_state assumption & d\_state=64 & d\_state=128 (for 1.3B) & 10K $\to$ 5{,}500 \\
Gate-then-norm ordering & norm($y$) $\cdot$ silu($z$) & norm($y \cdot$ silu($z$)) & 702 $\to$ 28.5 \\
\bottomrule
\end{tabular}
\end{table}

\section{Kernel Profiling}
\label{app:kernel}

Per-layer timing (GPU, batch=1), broken down by implementation source:

\begin{table}[h]
\centering
\caption{Per-layer kernel timing breakdown by implementation.}
\label{tab:kernel}
\begin{tabular}{@{}llrr@{}}
\toprule
Operation & Implementation & $\mu$s/layer & Fraction \\
\midrule
Scaffold matmul & cuBLAS INT8 (\texttt{torch.\_int\_mm}) & 47.1 & 23\% \\
LoRA projection & PyTorch (\texttt{torch.mm}) & 36.0 & 17\% \\
RMSNorm ($\times$2) & PyTorch & 22.5 & 11\% \\
Gate + Norm & PyTorch & 21.7 & 11\% \\
SSM step & Custom CUDA kernel & 6.7 & 3\% \\
Conv step & Custom CUDA kernel & 3.7 & 2\% \\
\bottomrule
\end{tabular}
\end{table}

cuBLAS and PyTorch library operations account for $\sim$62\% of per-layer compute. Custom CUDA kernels handle only the stateful operations (SSM recurrence and convolution shift register), comprising $\sim$5\% of per-layer time. The remaining $\sim$33\% is CUDA graph overhead, memory operations, and quantization/dequantization.

\section{Density Field Configuration}
\label{app:df}

For Mamba-2 with $d_{\text{state}} = 128$: $K = 23$ (field size $23 \times 23 = 529$), sigma-delta order 1, 8-bit fixed-point accumulator, DF-SSD applied at chunk boundaries (block\_len = 64).

\textbf{Universal rule:} $K^2 \approx 4 \times d_{\text{state}}$.

\section{Knowledge Atlas Category Details}
\label{app:atlas}

445 prompts across 19 categories: capitals (50), languages (30), continents (30), elements (30), physics (25), writers (25), animals (25), math (25), scientists (20), companies (20), currencies (20), food (20), music (20), sports (20), medical (20), history (20), colors (15), mythology (15), materials (15).

\end{document}